\documentclass[runningheads]{llncs}
\usepackage[T1]{fontenc}
\usepackage{amsmath}
\usepackage{graphicx}
\usepackage{booktabs}
\usepackage{tikz}
\usetikzlibrary{arrows.meta, positioning}

\usepackage[none]{hyphenat}
\begin{document}
\title{Agentopic: A Generative AI Agent Workflow for Explainable Topic Modeling}
%
%
\author{Brice Valentin Kok-Shun\inst{1}\orcidID{0000-0001-9923-5042} \and\newline 
Johnny Chan\inst{1}\orcidID{0000-0002-3535-4533} \and
Gabrielle Peko\inst{1}\orcidID{0000-0002-7878-3258} \and
David Sundaram\inst{1}\orcidID{0000-0003-1575-4174}}
\authorrunning{Kok-Shun et al.}
\institute{The University of Auckland, New Zealand}
\maketitle
\begin{abstract}
Agentopic is a novel agent-based workflow for explainable topic modeling 
that leverages the reasoning capabilities of Large Language Models (LLMs). 
Existing topic modeling approaches such as Latent Dirichlet Allocation (LDA) 
and BERTopic often lack transparency on how topics are assigned or grouped. 
Agentopic addresses this by using multiple agents that collaboratively 
perform topic identification, validation, hierarchical grouping, and natural language 
explanation. 
This design enables users to trace the reasoning behind topic assignments, 
enhancing interpretability without sacrificing accuracy. 
When seeded with topics from the British Broadcasting Corporation (BBC) dataset, 
Agentopic achieves an F1-score of 0.95, matching GPT-4.1, improving on LDA (0.93), 
and close to BERTopic (0.98). 
We used Agentopic to augment the BBC dataset with 
generated explanations to improve the dataset's richness and context. 
The unseeded Agentopic generated 2045 semantically coherent topics organized 
across six hierarchical levels, vastly enriching the original five-category structure. 
By embedding explainability throughout the workflow, Agentopic offers an interpretable 
alternative to black-box models, making it particularly valuable for crucial 
applications like finance and healthcare.

\end{abstract}
%
%
%

\section{Introduction}
Topic modeling plays a crucial role in organizing, summarizing, and extracting insights 
from large-scale textual corpora. From digital libraries and social media to customer 
feedback and policy documents, understanding the thematic structures hidden in 
unstructured text is essential for effective decision-making. However, as textual data 
becomes increasingly voluminous and complex, the demand for interpretable and adaptable 
topic modeling solutions has never been greater.

Traditional topic modeling approaches like Latent Dirichlet Allocation 
(LDA) \cite{blei_latent_2003}, have provided a statistical framework for discovering 
latent topics based on word co-occurrence patterns. Despite their wide adoption, these 
models often suffer from limited coherence, high sensitivity to hyperparameter tuning, 
and poor adaptability to domain-specific language. More recently, Large Language Model 
(LLM) methods like BERTopic \cite{grootendorst_bertopic_2022} have shown promise with 
transformer-based embeddings and clustering techniques. The advent of 
Generative AI (GenAI) and autonomous agents powered by LLMs 
introduces a new paradigm for topic modeling. Unlike statistical or 
embedding-based methods, GenAI agents can actively explore, refine, and detail 
topic structures through iterative prompt-based interactions \cite{hughes_ai_2025}. 

Explainability is critical in real-world applications of topic modeling;
especially in domains such as healthcare and finance where opaque 
clustering can lead to misinformed decisions or overlooked nuances. 
While models such as LDA and BERTopic produce semantically rich topic clusters, their 
reasoning often remains opaque and are dependent on post-hoc explanation 
techniques for justification \cite{rajendran_local_2024}. 

We propose Agentopic,
a novel agentic workflow that leverages the 
generative capabilities of GenAI to produce explainable and coherent topic 
models. By leveraging LLM-powered agents capable of 
topic suggestion, coherence evaluation, and explanation generation, we address the 
challenges of topic quality and interpretability. We draw from Information Systems' 
qualitative research process of thematic coding \cite{myers_qualitative_1997} to 
inform the design of a multi-agent workflow that systematically identifies, validates, 
and groups topics while providing human-readable explanations for each stage of the 
process.

We evaluate Agentopic on the British Broadcasting Corporation (BBC) dataset 
\cite{greene_practical_2006}. Our experiments 
also include comparisons with traditional topic modeling approaches such as LDA and
BERTopic, as well as LLM-based models like GPT-4.1. The results demonstrate that Agentopic's 
performance is on par with all the models used for benchmarking. It improves on
current models by enhancing transparency and trust through its explanation mechanism 
which we used to add an explainability layer to the BBC dataset. 
Our work bridges the gap between statistical rigor, semantic richness, and human 
interpretability in topic modeling.

The remainder of this paper is organized as follows. Section~2 reviews related work in 
topic modeling, agentic and GenAI approaches, and explainability. Section~3 details 
the methodology, including the dataset, text preprocessing, and the experimental
Agentopic workflow. 
Section~4 presents experimental results, highlighting quantitative 
and qualitative findings. Section~5 discusses the implications, strengths, and 
limitations of our approach, and outlines directions for future research. 
Finally, Section~6 concludes the paper by summarizing our main contributions and 
the significance of explainable topic modeling.

\section{Related Works}
This section reviews the literature relevant to our work. We discuss topic modeling
techniques, the emergence of agentic and GenAI approaches, and the role of 
explainability in enhancing the interpretability of topic models and enriching datasets.

\subsection{Topic Modeling Techniques}
Topic modeling approaches can be grouped into four main categories: algebraic, 
fuzzy, probabilistic, and neural \cite{abdelrazek_topic_2023}. One of the most 
influential works in the probabilistic category is the LDA \cite{blei_latent_2003}, 
which introduced a generative probabilistic model for discovering latent 
topics in text corpora. LDA has been widely adopted due to its ability to capture 
complex topic structures. The algebraic, fuzzy, and probabilistic approaches
work well on well-edited, long documents with consistent word co-occurrence patterns, 
but struggle with short, sparse, and fast-changing texts like social media posts. 
Challenges in such contexts include limited text length, vocabulary sparsity, high 
volume, and rapid topical shifts \cite{churchill_evolution_2022}.

Tasks such as stopword removal, stemming, and lemmatization are critical for topic 
modeling and need to be tailored for different languages. These preprocessing steps 
greatly influence the clarity and efficiency of the output. Omitting them can lead 
to an unnecessarily large vocabulary, increasing computational costs; 
over-preprocessing the text can reduce the semantic quality of the resulting 
topics~\cite{chauhan_topic_2022}.

The introduction of deep learning and transformer-based models has enhanced the 
semantic richness of topic representations. These models rely on text 
embeddings to capture contextual relationships between words, allowing for coherent 
topic discovery. Examples of such embeddings include GloVe \cite{pennington_glove_2014}, 
Word2Vec \cite{mikolov_efficient_2013}, and 
transformer-based models like BERT \cite{devlin_bert_2018}. These embeddings are used 
in models like BERTopic \cite{grootendorst_bertopic_2022}, which combines BERT-based 
embeddings with clustering techniques to produce semantically richer topic clusters.
Other models may opt to derive their own embeddings to capture domain-specific nuances
and connections to topics of interest \cite{taggu_deep_2024}.

While topic modeling can be done in an unsupervised manner, creating a topic model that
can be reused requires training an algorithm on labeled data. These labels allow it to
learn the intricate patterns essential for language understanding and generation 
\cite{paullada_data_2021,singla_overcoming_2019}. Academics and practitioners have to go 
through the painstaking process of creating labeled datasets, which involves either 
manually assigning topics to all records in a dataset, reviewing the topic model's 
output and manually correcting labels to each topic based on its content in a 
semi-automated manner. An alternative is obtaining a suitable existing labeled dataset,
however, this poses a significant challenge due to data quality and scarcity issues 
\cite{agarwal_editorial_2014,bansal_systematic_2022}.

The semi-automated approach is often preferred, as it allows for a more efficient
creation of topic models. However, it still requires significant human effort to 
ensure the quality and accuracy of the labels. This process can be time-consuming and
labor-intensive, particularly for large datasets. A potential solution to this
problem is the use of GenAI to aid in the topic labeling process,
which can significantly reduce the time and effort required to create high-quality
topic models.

\subsection{Agents, Generative AI and Explainability}
The emergence of GenAI has revolutionized the field of natural language
processing by enabling the development of LLMs that can generate human-like text, 
understand context, and perform complex explanatory tasks. Models, such as OpenAI's GPT 
series, have demonstrated remarkable capabilities in various natural language processing
tasks, including 
text generation, summarization, translation, and question answering \cite{mohammad_large_2023}.
Recently, the rise of agentic systems has further enhanced the capabilities of GenAI by
introducing autonomous agents that can perform tasks, make decisions, and interact with
users or act independently \cite{hughes_ai_2025}. Agents can act as teammates and 
collaborate and assist humans in various tasks to improve accuracy and
efficiency \cite{dennis_ai_2023,hughes_ai_2025}.

The use of GenAI in topic modeling allows for the generation of more coherent and
contextually relevant topics, as these models can leverage vast amounts of pre-trained
knowledge. Coupled with agentic systems, GenAI further facilitates the identification of
topics in texts by supporting the users in generating topics, validating, and
grouping topics. While the accuracy of the GenAI models is beneficial, there has been a
call for more explainable AI (XAI) systems that can provide insights into the reasoning 
behind the model's outputs rather than accepting the black-box system's results
\cite{rajendran_local_2024,xie_knowledge_enhanced_2024}. This is particularly important 
in applications where the consequences of misinterpretation or misclassification can be 
significant \cite{van_der_velden_explainable_2022}. However, XAI should not only be
restricted to critical fields such as healthcare, finance, and public policy; the 
benefits of XAI can be extended to any domain where explainability and interpretability
can support better decision-making and user trust in AI systems. For instance,
XAI can be used to improve reproducibility and transparency in scientific research simply
by supporting researchers in producing clear explanations of how a dataset was labeled.
While this is a critical aspect of scientific research, it is often overlooked as it is
time consuming and labor-intensive to document the justification for each label. GenAI
can assist researchers in generating these explanations, thereby enhancing the
reproducibility and transparency of their work \cite{ries_reproducibility_2024}.

A method used in topic modeling to improve explainability is the use of Graph Neural 
Topic Models (GNTMs) \cite{zhu_graph_2023} that create a graphical representation of 
the corpus and creates document-topic and topic-word probability distributions. 
These models are still limited in their explanatory and interpretive power. More 
human-friendly approaches such as decision trees or simpler probabilistic models such as 
Naïve Bayes have been proposed as potential solutions \cite{rajendran_local_2024}.
These methods still require an understanding of the underlying mathematics 
for interpretation, which can be a barrier for many users. The use of GenAI in providing
natural language explanations for topic models can help bridge this gap by providing
human-readable explanations that are easy to understand and interpret. Such explanations 
are preferred due to their human-friendly nature over mathematical or statistical 
explanations \cite{mittelstadt_explaining_2019,camburu_e_snli_2018}.

\section{Methodology}
In this section, we detail the methodology underlying Agentopic, our proposed agentic 
workflow for explainable topic modeling. We outline the Agentopic workflow, agentic 
system design, and use of GenAI for topic discovery and explanation. This 
section also covers the dataset, models, and metrics used for evaluation.

\subsection{BBC Dataset}
To evaluate the performance and explainability of our proposed Agentopic workflow, 
we utilize the publicly available BBC dataset \cite{greene_practical_2006}. 
This dataset consists of 2225 news 
articles collected from the BBC News website, covering a variety of stories published 
between 2004 and 2005. Each document is labeled with one of five high-level topical 
categories: \textit{Business}, \textit{Entertainment}, \textit{Politics}, \textit{Sport},
and \textit{Tech}. These categories provide a ground truth for validating the thematic 
coherence and accuracy of topic modeling outputs. The BBC News dataset 
is widely used in text classification and clustering research due to its clean structure, 
well-defined class labels, and moderate size, making it suitable for both qualitative 
and quantitative evaluation.

To prepare the dataset, we implemented a text preprocessing pipeline 
to enhance semantic clarity and reduce lexical noise. The \textit{Title} field 
was not preprocessed as it is short text whose semantic value will be diluted if 
preprocessed. The \textit{Description} field was preprocessed 
because it is significantly longer and contains complex linguistic structures 
that can negatively impact topic modeling by introducing noise if not properly cleaned. 
The preprocessing steps are as follows:
\begin{enumerate}
    \item \textbf{Contraction Expansion}: Contracted forms were expanded to their full 
            versions using the \texttt{contractions} Python library.
    \item \textbf{Lowercasing}: This ensures consistent word frequency counts.
    \item \textbf{Punctuation and Digit Removal}: Special characters and numeric 
            digits were removed using regular expressions to reduce syntactic noise.
    \item \textbf{Whitespace}: Extra spaces were removed for consistent token separation.
    \item \textbf{Stopword Removal}: Standard English stopwords from NLTK were removed. 
    \item \textbf{Lemmatization}: Verbs were lemmatized using the WordNet lemmatizer to 
            preserve the contextual integrity of words, compared to stemming.
\end{enumerate}

This multi-step pipeline significantly reduced noise and vocabulary size, ensuring that 
the resulting textual features are both computationally manageable and semantically 
informative for downstream topic modeling.

\subsection{Agentopic Workflow}
Agentopic is a multi-agent system that orchestrates specialized generative agents 
in a coordinated pipeline to perform explainable topic modeling. Each agent is 
designed to handle a specific cognitive or analytic task-such as topic discovery, 
validation, grouping, or hierarchy construction-and communicates with others through 
a structured feedback loop.

\begin{center}
\resizebox{\textwidth}{!}{
\begin{tikzpicture}[
    box/.style={draw, rounded corners, text width=2.8cm, align=center, minimum height=1.1cm, font=\Large}, 
    ->, >=Stealth
]

\node[box] (start) {\textbf{Start}\\Input Texts};
\node[box, right=of start, text width=4cm] (A) {\textbf{Identification}\\Generative or seeded};
\node[box, right=of A, yshift=2.4cm] (B) {\textbf{Grouping}\\Combine topics};
\node[box, below=of B, yshift=-2.4cm, text width=3.5cm] (C) {\textbf{Review}\\Validate topics};
\node[box, right=of C, text width=3.5cm] (end1) {\textbf{End}\\Seeded topics};
\node[box, right=of B] (D) {\textbf{Review}\\Validate groups};
\node[box, right=of D, text width=3.5cm] (E) {\textbf{Hierarchy}\\Hierarchical grouping};
\node[box, right=of E] (end2) {\textbf{End}\\Topic Hierarchy};
\node[box, right=of B, yshift=-2.7cm] (F) {\textbf{Vector DB}\\Embeddings};

\draw[very thick,->] (start) -- (A);
\draw[very thick,->] (A) to[bend left=20] (C); 
\draw[very thick,->] (C) to[bend left=20] node[left,pos=0.5,yshift=-6pt]{\Large Fix errors} (A); 
\draw[very thick,->] (C) -- (B);
\draw[very thick,->] (C) -- (end1);
\draw[very thick,->] (B) -- (D);
\draw[very thick,->] (D) to[bend right=40] node[above,pos=1.2,yshift=20pt]{\Large Fix errors} (B); 
\draw[very thick,->] (D) -- (E);
\draw[very thick,->] (E) to[bend right=30] node[above,pos=0.4,yshift=4pt]{\Large Refinement iterations} (B); 
\draw[very thick,->] (E) -- (end2);
\draw[very thick,dashed,<->] (C) to[bend left=30] (F); 
\draw[very thick,dashed,<->] (D) to (F); 

\end{tikzpicture}
}
\end{center}
\vspace{-2.5em}
\begin{figure}[h]
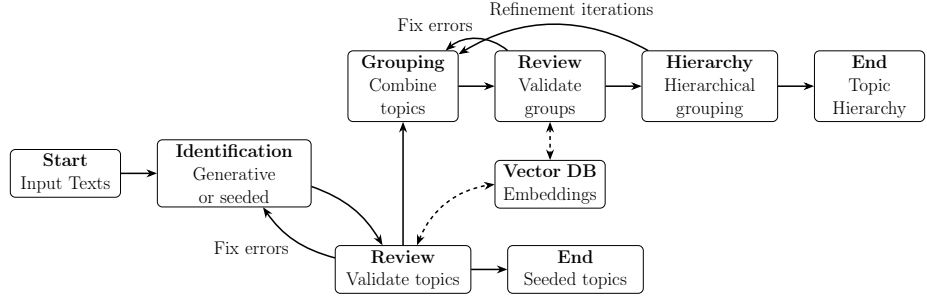

    \centering
    \begin{minipage}{\textwidth}
        \centering
        \caption{Agentopic workflow: multi-agent orchestration for explainable topic modeling.}
        \label{fig:architecture}
    \end{minipage}
\end{figure}
\vspace{-1em}

The agentic workflow depicted in Figure~\ref{fig:architecture} is inspired by the 
qualitative research process of \cite{myers_qualitative_1997} which describes a 
process of thematic coding for topic identification.
The workflow is also inspired by GNTMs \cite{zhu_graph_2023} which create topic 
representations. The Agentopic workflow utilizes text embeddings of the topics and
explanations generated for both review and hierarchical grouping. The workflow 
consists of the following agents:

\begin{itemize}
    \item \textbf{Topic Identification Agent}: Initiates the process by identifying 
            candidate topics from raw text or user-provided seeds. This agent applies 
            the \textit{data familiarization} and \textit{initial coding} steps. 
            Explanations are generated to justify the identified topics.
    \item \textbf{Topic Review Agent}: Evaluates the generated topics for relevance and 
            proper formatting. If seeded topics are provided, the topics are 
            checked against the seeds, the workflow ends and topics and explanations 
            are returned. It passes the errors back to the Topic Identification Agent 
            if corrections are needed. The topic names and explanations are embedded and 
            stored in a vector database.
    \item \textbf{Topic Grouping Agent}: Groups validated topics into broader thematic 
            groups following the \textit{thematic grouping} step. This agent ensures 
            that related topics are grouped together. The groups are generated based on the 
            topic names and explanations from the Topic Identification Agent. 
            This agent also generates a group description for each thematic group. 
            In subsequent iterations, it inherits the hierarchy from the previous iteration.
    \item \textbf{Group Review Agent}: This agent executes the \textit{theme refinement} step. 
            The groups are examined for structural validity and missing topics. 
            It communicates with the Topic Grouping Agent for corrections if needed; 
            otherwise, advances to hierarchy construction. It also embeds the group name and
            description and stores it in the vector database.
    \item \textbf{Hierarchy Construction Agent}: Imposes a hierarchical structure 
            on grouped topics, identifying general-to-specific relationships. 
            It synthesizes topic trees for simpler \textit{thematic interpretation}.
            In multiple iterations, it refines the hierarchy feeds back into the Topic 
            Grouping Agent for further grouping and refinement. At the end of the workflow,
            it returns the final topic hierarchy.
\end{itemize}

This agent-based design enables modularity, accountability, and transparency at each 
stage of topic modeling. By delegating distinct roles to specialized agents, Agentopic 
not only improves coherence and structure but also ensures that each step is explainable 
by design. Unlike monolithic LLM-based models or black-box clustering pipelines, this 
architecture supports traceable decision-making, and extensible control over how topics 
are identified, grouped, and interpreted.

\subsection{Experimental Setup - Seeded Topic Identification}
To assess the effectiveness of the proposed Agentopic workflow, we conducted 
experiments on the BBC News dataset using three modeling paradigms: (1) statistical 
topic modeling via LDA with TF-IDF vectorizer, (2) embedding-based topic 
modeling using BERTopic, and (3) GenAI-based topic inference using GPT-4.1 
variants, both standalone and within the Agentopic multi-agent architecture. LDA and 
BERTopic models were trained using the labeled BBC dataset. No fine-tuning was done 
for the GPT variants. Our goal was to compare topic identification performance across 
these approaches while assessing the added value of agentic orchestration.\\

\textbf{Data Preparation:} Different preprocessing pipelines were applied depending on 
the modeling paradigm. For GPT-4.1 (all variants), Agentopic, and BERTopic,
the models received raw, unprocessed text from the \textit{Title} and \textit{Description} 
field of each article. This design reflects the capacity of LLMs to operate 
directly on unstructured text using prompt engineering. The preprocessed dataset was 
used for LDA only.\\

\textbf{Model Configurations:} The following configurations were used for each
model:
\begin{itemize}
    \item \textbf{LDA}: Applied using a TF-IDF vectorizer. The dominant topic for each 
            document was mapped to one of the five predefined labels for classification 
            evaluation.
    \item \textbf{BERTopic}: Utilized the all-MiniLM-L6-v2 sentence embedding model to 
            generate dense vector representations of documents. A logistic regression 
            classifier was trained on the topic-document pairs to perform category prediction.
    \item \textbf{GPT-4.1 Variants (nano, mini, and full)}: Prompted directly with article 
            titles and descriptions and tasked with assigning the most likely topic label.
    \item \textbf{Agentopic}: Leveraged the same GPT-4.1 variants, but operated within a 
            multi-agent workflow, where specialized agents performed topic identification, 
            review, grouping, and hierarchical reasoning to generate an interpretable and 
            validated topic label.
\end{itemize}

\textbf{Evaluation:} All models were evaluated on their ability to classify documents 
into the five ground truth categories of the BBC dataset. Precision, recall, and 
F1-score were computed for each category and averaged to obtain overall performance metrics. 
The metrics are defined as:
\begin{align}
    Precision &= \frac{1}{N} \sum_{i=1}^{N} \frac{TP_i}{TP_i + FP_i} \\
    Recall    &= \frac{1}{N} \sum_{i=1}^{N} \frac{TP_i}{TP_i + FN_i} \\
    F1_{\text{score}} &= \frac{1}{N} \sum_{i=1}^{N} \frac{2 \cdot Precision_i \cdot Recall_i}{Precision_i + Recall_i}
\end{align}
where $N$ is the number of topics, $TP_i$ is the number of true positives, $FP_i$ is the number 
of false positives, and $FN_i$ is the number of false negatives for topic $i$. 
$F1_{\text{score}}$ is the harmonic mean of precision and recall, providing a balanced measure 
of overall performance.

\subsection{Experimental Setup - Generative Topic Identification}
To evaluate the generative capabilities of Agentopic, we conducted experiments using the 
BBC dataset, emphasizing the workflow's ability to generate coherent and 
relevant topics. The evaluation focused on two central dimensions: \textit{coherence}, 
referring to the internal semantic consistency of each topic, and \textit{coverage}, 
indicating how comprehensively the model's topics span the thematic landscape of the 
dataset. These dimensions are crucial for assessing the quality of the outputs \cite{churchill_evolution_2022}.

Agentopic was prompted using the \textit{Title} and \textit{Description} fields of 
each article and employed its agentic workflow to refine the generated topics. 
Coherence was assessed by examining whether the topics formed a semantically meaningful 
and interpretable group, minimizing noise or unrelated terms. 
Coverage was judged by comparing the generated topic set to the original five BBC 
categories, observing whether the generated topics collectively captured the breadth of 
themes present in the corpus.

\section{Results}
In this section, we present the results of our experiments, evaluation metrics, and 
provide both quantitative and qualitative analyses of our approach.

\subsection{Seeded Topic Identification}
Our results demonstrate that Agentopic achieves excellent topic identification 
performance while offering a substantial advantage in explainability and 
interpretability over traditional and LLM-based baselines. Table~\ref{tab:bbc_accuracy} 
presents per-category and overall accuracies across different models on the BBC dataset.
\begin{table}[h]
\centering
\caption{Topic Identification F1-Score by Model and Category on the BBC Dataset}
\begin{tabular}{lcccccc}
\toprule
\textbf{Model} & \textbf{Business} & \textbf{Entertainment} & \textbf{Politics} & \textbf{Sport} & \textbf{Tech} & \textbf{F1-Score} \\
\midrule
\multicolumn{7}{l}{\textbf{Baselines}} \\
gpt-4.1-nano  & 0.87 & 0.90 & 0.87 & 0.98 & 0.80 & 0.88 \\
gpt-4.1-mini  & 0.90 & 0.93 & 0.94 & 0.99 & 0.86 & 0.92 \\
gpt-4.1       & 0.93 & 0.95 & 0.93 & 0.99 & 0.94 & 0.95 \\
LDA           & 0.91 & 0.94 & 0.89 & 0.98 & 0.92 & 0.93 \\
BERTopic      & 0.97 & 0.98 & 0.97 & 0.99 & 0.97 & 0.98 \\
\midrule
\multicolumn{7}{l}{\textbf{Agentopic}} \\
gpt-4.1-nano  & 0.82 & 0.92 & 0.83 & 0.98 & 0.81 & 0.87 \\
gpt-4.1-mini  & 0.89 & 0.93 & 0.94 & 0.99 & 0.87 & 0.92 \\
gpt-4.1       & 0.94 & 0.95 & 0.93 & 0.99 & 0.94 & 0.95 \\
\bottomrule
\end{tabular}
\label{tab:bbc_accuracy}
\end{table}

BERTopic achieves the highest overall F1-score (0.98), marginally
outperforming all other models. This performance is expected given that BERTopic 
incorporates supervised classification via logistic regression on rich semantic 
embeddings, allowing it to fine-tune boundaries between categories. However, this 
comes at the cost of black-box interpretability, as the clustering decisions and 
classifier boundaries are not easily explainable.

Agentopic, by contrast, achieves parity with or exceeds the performance of raw 
GPT-4.1 models across all variants. The agentic version of GPT-4.1-mini matches 
its non-agentic counterpart (0.92), and Agentopic with full GPT-4.1 maintains 
peak F1-score at 0.95 overall-demonstrating that agentic orchestration does 
not degrade performance, even while introducing structured reasoning and explanations.

\subsection{Generative Topic Identification}
\begin{figure}[h]
    \centering
    \includegraphics[width=\textwidth]{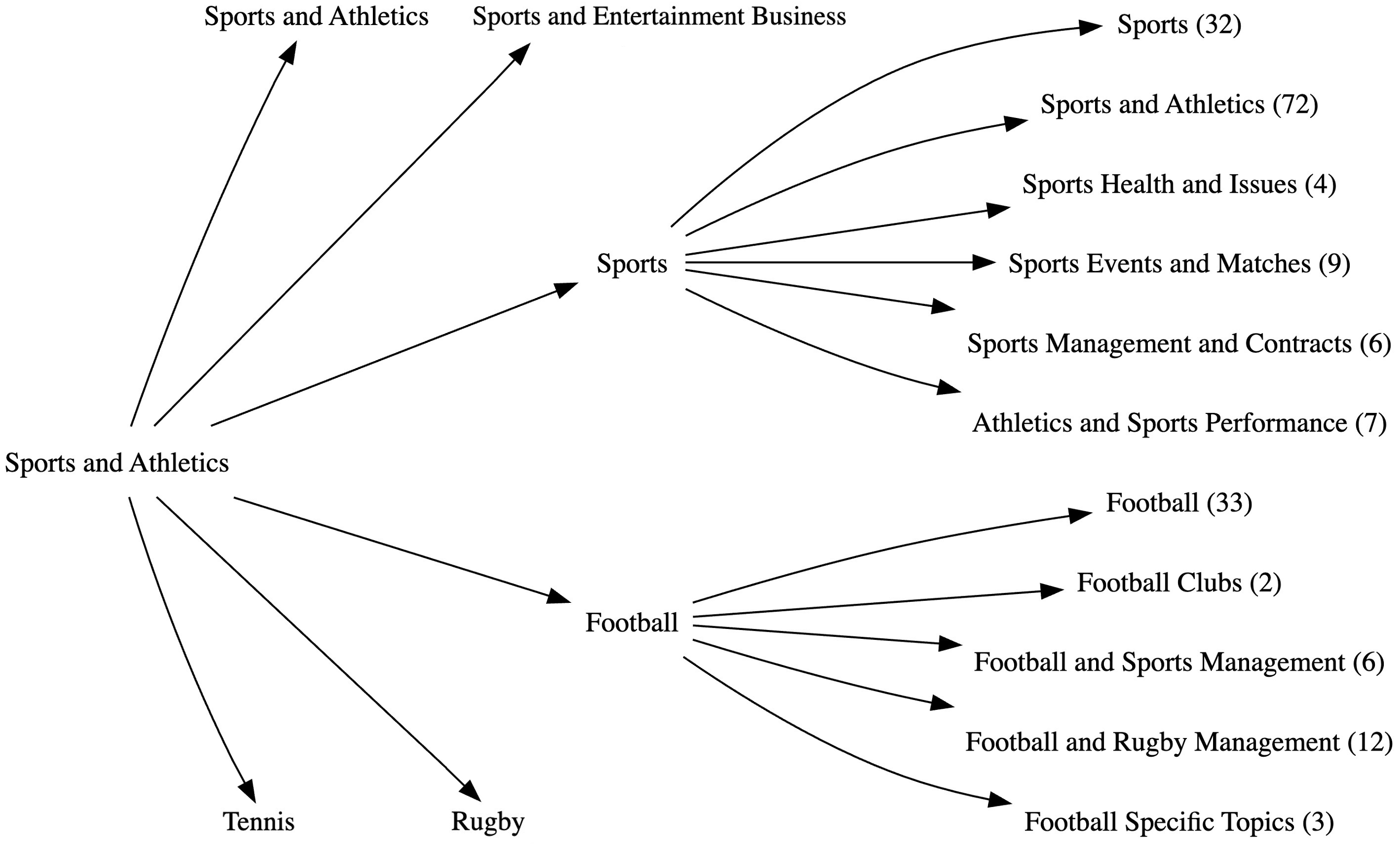}
    \caption{Section of the Agentopic hierarchy showing sports-related topics.}
    \label{fig:sports_example}
\end{figure}

Our experiments also included a qualitative analysis of the generated topics. 
We found that Agentopic was able to produce coherent and contextually relevant 
topics that aligned well with user expectations. Figure~\ref{fig:sports_example} 
illustrates an excerpt from the topic hierarchy produced by Agentopic for a 
sports-related news article.

The Agentopic workflow generated a total of 2045 topics, organized hierarchically 
across six levels, with four top-level groups: 
\textit{Entertainment and Media}, \textit{Industry and Environment}, 
\textit{Legal and Regulatory}, and \textit{Sports and Events}. This hierarchical 
design contrasts with the flat 5 categories of the BBC 
dataset (\textit{Business}, \textit{Entertainment}, \textit{Politics}, 
\textit{Sport}, \textit{Tech}). Notably, most Tech articles were distributed 
across the other four categories based on their contextual content, suggesting 
Agentopic offers a more semantically grounded classification.

In terms of coverage, Agentopic successfully recovers and expands upon the 
topic space of the original dataset. It uncovers a broader spectrum of subtopics 
that reflect diverse subject areas within each category. This expanded coverage 
is particularly evident in the fine-grained segmentation of each domain. 
For example, \textit{Entertainment} is broken down into 16 branches, including 
film, music, television, and celebrity culture. \textit{Politics} is separated 
into domestic affairs, international relations, elections, and security. 
\textit{Business} spans legal, regulatory, financial, and industrial sectors, 
and \textit{Sport} is detailed by specific athletic disciplines, health aspects, 
and sports management. Such granularity reveals topic dimensions overlooked 
by non-hierarchical models.

Regarding coherence, the generated topics consistently grouped semantically 
similar and topically relevant terms. Manual inspection showed that subtopics 
within a branch exhibited strong internal consistency alignment, 
supporting interpretability. For example, subtopics under \textit{Sports and Events} 
clustered terms around training, competition, and event outcomes without introducing 
noisy or irrelevant vocabulary. Similarly, branches under \textit{Legal and Regulatory} 
maintained focus on policy, governance, and institutional processes.

\subsection{Explainability and Interpretability}
Unlike traditional topic models, which often output opaque clusters or require 
post-hoc interpretation, Agentopic generates explanations for each topic identified
in the texts, descriptions of the topic groups, and hierarchical topic model. These 
explanations and descriptions are in natural language, making them accessible
to users without requiring deep technical expertise.
The provided examples are from the BBC dataset we augmented with explanations.
They show how both seeded and generated topics are aligned.

\begin{quote}
    \textit{``Ad sales boost Time Warner profit - Quarterly profits at US 
    media giant TimeWarner jumped 76\% to \$1.13bn (£600m) for the three 
    months [...] TimeWarner is to restate its accounts as part of efforts 
    to resolve an inquiry into AOL by US market regulators [...]''}

    \textbf{Topics:} Business (Seeded); Economy and Finance (Generated)
    \\\textbf{Explanation:} The text discusses Time Warner's financial 
    performance, profits, sales, investments, and business activities, 
    all are related to Business.
    \\\textbf{Generated Group Description:} These topics cover taxation, 
    economic policies, financial services, corporate legal and financial problems, 
    and investment. This group includes economic conditions, finance, and
    currency exchange topics.
\end{quote}

In the example above, the model assigns the topic \textit{Economy and Finance}
which is related to the original \textit{Business} category. The explanation 
describes how the text discusses financial performance, profit figures, and 
regulatory context. This not only validates the topic assignment but also 
instills confidence in users reviewing the model's outputs in high-stakes 
domains like corporate finance or media regulation.

\begin{quote}
    \textit{``Claxton hunting first major medal - British hurdler Sarah Claxton 
    is confident she can win her first major medal at next month [...] 
    Claxton will see if her new training regime pays dividends at the European 
    Indoors [...]''}

    \textbf{Topics:} Sport (Seeded); Sports (Generated)
    \\\textbf{Explanation:} The text discusses British hurdler Sarah Claxton's 
    performance, training, and prospects at the European Indoor Championship, which 
    are all related to athletics and sports competitions.
    \\\textbf{Generated Group Description:} This group includes topics related to
    various sports, sports events, injuries, and management. This group includes topics
    related to sports events, management, competitions, and football matches.
\end{quote}

Similarly, the model identifies the above article's domain as \textit{Sport} 
(same as original category), explicitly linking content about training, competition, 
and athletic performance to its topic hierarchy. The generated group description 
adds value by situating the example within a broader conceptual cluster-sports 
events, injuries, and management-highlighting how Agentopic contextualizes specific 
articles within wider discourses.

By outputting topic labels and contextual explanations, it enhances 
interpretability, builds user trust, and supports use in domains where 
traceability and accountability are critical. 

\section{Discussion}
This section discusses the implications of our findings, analyzes the strengths and 
weaknesses of the Agentopic workflow, and explores its broader impact on the field 
of explainable topic modeling. We also highlight key areas for future research and 
practical considerations for deploying agentic topic modeling systems in 
real-world applications.

\subsection{Comparison with Existing Methods}
Overall, while BERTopic sets the upper bound for F1-score, it lacks any form of 
native topic explanation or semantic traceability. In contrast, Agentopic delivers 
near-optimal classification while providing interpretable outputs, including named 
topics, justifications, and hierarchical context. These properties are critical in 
applied domains such as healthcare where transparency is as important as 
predictive performance \cite{van_der_velden_explainable_2022}.

When examining the per-category performance, we observe that Agentopic maintains
high F1-scores across all categories, with notable strengths in \textit{Entertainment} and
\textit{Sport}, where it achieves 0.92 and 0.98 respectively. These categories are lexically
rich and topically distinct, making them easier to classify even with simpler
baselines. In contrast, Agentopic shows a slight dip in \textit{Business} and 
\textit{Politics}, where the complexity of themes and potential overlaps with other 
categories introduce challenges. For example, Agentopic-GPT-4.1-nano achieves 0.82 in
\textit{Business} and 0.83 in \textit{Politics}, compared to 0.87 for the non-agentic
GPT-4.1-nano. This suggests that while Agentopic's multi-step workflow is beneficial,
it may introduce some overhead in categories where the topics are closely related or
ambiguous.

Agentopic demonstrates a high degree of granularity by generating 2045 topics, 
far surpassing the original five-category structure of the BBC dataset. This level of 
detail enhances coverage, as the workflow captures a significantly broader range of 
concepts, enabling richer thematic representation across articles, a key goal in modern 
topic modeling tasks \cite{blei_probabilistic_2012}. The six-level hierarchical 
taxonomy allows users to navigate between coarse and fine levels of abstraction, ranging 
from general categories like \textit{Digital Media} to highly specific subtopics such as 
\textit{Music and Digital Media} or \textit{Cybersecurity and Spam}. Such multi-level 
representations are vital in tasks requiring nuanced textual 
understanding, such as policy analysis or information retrieval \cite{dieng_topic_2020}. 
This hierarchical structure contributes to both coverage and coherence: it organizes 
related subtopics under meaningful groups, ensuring topical breadth without 
sacrificing thematic integrity \cite{roder_exploring_2015}.

Additionally, Agentopic excels in coherence, with each topic node displaying strong 
internal consistency and semantic clarity. Natural language explanations and descriptions, 
contextual alignment, and well-structured parent-child relationships enhance 
explainability. 

\subsection{Explainability and Topic Modeling}

Traditional statistical models like LDA rely on 
probability distributions over words and documents, offering little interpretability 
beyond top ngrams. While newer models such as BERTopic leverage contextual 
embeddings to produce more coherent topics, they still operate as black-boxes with 
limited transparency regarding how and why specific topic assignments are made.
Their lack of explainability presents serious limitations. Users are 
often left to manually interpret clusters, and cannot trace the rationale behind 
topic assignments or groupings. This becomes problematic in high-stakes scenarios 
where model outputs influence human decisions, and where accountability, justification, 
and traceability are essential \cite{rajendran_local_2024,xie_knowledge_enhanced_2024,ries_reproducibility_2024}.

Agentopic addresses this gap by embedding explainability into the core architecture 
of the modeling process. Rather than treating topic discovery as a one-shot statistical 
or embedding-based task, Agentopic operationalizes it as a sequence of agent-driven 
analytical steps. Each agent in the pipeline not only performs a specialized function, 
but also generates natural language explanations justifying its actions. 
These explanations are clear, human-readable rationales for why articles are associated 
with a given topic or why certain topics are grouped together. Such explanations 
make it easier to understand the rationale behind the grouping and are more digestible to
individuals with varying levels of technical expertise. This makes them more accessible to
the general public \cite{mittelstadt_explaining_2019,camburu_e_snli_2018}. 
For example, when assigning a document to a topic like Economy and Finance, the system 
explicitly identifies the relevant financial terms, context, and thematic links in the 
document. Additionally, each generated topic is placed within a broader semantic hierarchy, 
providing structural explainability through parent-child relationships. Agentopic also 
provides a visual representation of the topic hierarchy, allowing users to see how topics
are related and how they fit into the overall structure. This visual representation further
enhances the interpretability of the model's outputs, making it easier for users to
navigate the topic space and understand the relationships between topics 
\cite{alexander_serendip_2014}.

This design allows users to audit the model's logic, understand the topic groupings, 
and identify ambiguous or incorrect groupings. It reduces 
the gap between interpretability and usability, and makes topic modeling 
more aligned with human cognitive processes via natural language explanations and 
hierarchical topic grouping visualizations. Agentopic contributes to topic modelling 
techniques by providing a transparent, agentic framework that enhances the
explainability of topic discovery. This also highlights the importance of auditability 
and explainability in real-world applications where interpretability is as important 
as accuracy.

\subsection{Limitations and Future Work}
Agentopic currently assumes that the generative agent's 
reasoning is always contextually accurate, yet there is no mechanism for quantitative 
coherence validation or human-in-the-loop correction. Misclassifications may be masked 
by plausible-sounding explanations. User feedback loops and coherence scoring would 
increase reliability and control of the workflow. These improvements 
would further reinforce Agentopic's potential as a robust, interpretable, and adaptive 
workflow for explainable topic modeling.

Our qualitative analysis of the topic hierarchy also uncovered some topic grouping
inconsistencies. For instance, in Figure~\ref{fig:sports_example}, \textit{Sports and Athletics} 
was a subgroup under \textit{Sports and Athletics}, which is redundant and could lead to
confusion. Another example specific to the BBC dataset are the \textit{Tech} topics that
were distributed across the other four categories, which could lead to a lack of
clarity in the topic hierarchy. This may potentially be due to the article's content in 
the dataset; this is difficult to determine as the topic categories are not justified, 
further highlighting the need for explainability in topic modeling. These inconsistencies
highlight a potential area for improvement in the Agentopic workflow. It may benefit
from additional refinement steps to ensure that the hierarchy is not only semantically
coherent but also structurally sound. Future work could explore automated methods for
merging redundant topics such as pruning the topic hierarchy tree.

While Agentopic performs well on English texts, its 
multilingual capabilities remain untested despite GPT models being multilingual;
in addition to further experiments on non-English datasets, future work should explore
Agentopic's performance on other data sources such as social media, scientific journals,
or legal documents.

The limitations and future work outlined highlight the need for
continuous refinement and enhancement of the Agentopic workflow. By addressing these
issues, we can further improve the robustness, reliability, and applicability of Agentopic
in various domains.

\section{Conclusion}
This paper presents Agentopic, a multi-agent workflow for explainable topic modeling that 
combines the generative power of LLMs with structured topic 
discovery. Unlike traditional models such as LDA or BERTopic, Agentopic generates 
interpretable, hierarchical topic structures with natural language explanations at 
each stage.

Experiments on the BBC dataset show that Agentopic achieves strong performance with 
an overall F1-score of 0.95; this is the same as GPT-4.1, higher than LDA (0.93)
and closely trailing BERTopic's 0.98, while offering significantly improved interpretability 
through explanations of why a particular topic was chosen, which we used to augment 
the existing dataset. 
Agentopic generated 2045 fine-grained topics across a six-level hierarchy, 
capturing rich semantic distinctions beyond the dataset's original five labels.

By integrating explainability directly into its workflow, Agentopic offers a 
transparent and extensible alternative to black-box topic modeling methods. 
This inherent explainability makes Agentopic particularly well-suited for 
sensitive or high-stakes domains where transparency and traceability are 
essential. By elevating topic modeling from a black-box task to an agentic, 
interpretable process, Agentopic contributes to trustworthy, 
human-aligned topic modeling systems. Future work will explore multilingual capabilities 
and user-in-the-loop refinement to further enhance its transparency and adaptability.

%
%
\bibliographystyle{splncs04}
\bibliography{arxiv}

\begin{thebibliography}{10}
\providecommand{\url}[1]{\texttt{#1}}
\providecommand{\urlprefix}{URL }
\providecommand{\doi}[1]{https://doi.org/#1}

\bibitem{abdelrazek_topic_2023}
Abdelrazek, A., Eid, Y., Gawish, E., Medhat, W., Hassan, A.: Topic modeling
  algorithms and applications: {A} survey. Information Systems  \textbf{112},
  102131 (Feb 2023)

\bibitem{agarwal_editorial_2014}
Agarwal, R., Dhar, V.: Editorial —{Big} {Data}, {Data} {Science}, and
  {Analytics}: {The} {Opportunity} and {Challenge} for {IS} {Research}.
  Information Systems Research  \textbf{25}(3),  443--448 (Sep 2014)

\bibitem{alexander_serendip_2014}
Alexander, E., Kohlmann, J., Valenza, R., Witmore, M., Gleicher, M.: Serendip:
  {Topic} model-driven visual exploration of text corpora. In: 2014 {IEEE}
  {Conference} on {Visual} {Analytics} {Science} and {Technology} ({VAST}). pp.
  173--182. IEEE, Paris, France (Oct 2014)

\bibitem{bansal_systematic_2022}
Bansal, A., Sharma, R., Kathuria, M.: A {Systematic} {Review} on {Data}
  {Scarcity} {Problem} in {Deep} {Learning}: {Solution} and {Applications}. ACM
  Computing Surveys  \textbf{54}(10s),  1--29 (Jan 2022)

\bibitem{blei_probabilistic_2012}
Blei, D.M.: Probabilistic topic models. Communications of the ACM
  \textbf{55}(4),  77--84 (Apr 2012)

\bibitem{blei_latent_2003}
Blei, D.M., Ng, A.Y., Jordan, M.I.: Latent dirichlet allocation. J. Mach.
  Learn. Res.  \textbf{3},  993–1022 (Mar 2003)

\bibitem{camburu_e_snli_2018}
Camburu, O.M., Rocktäschel, T., Lukasiewicz, T., Blunsom, P.: e-{SNLI}:
  {Natural} {Language} {Inference} with {Natural} {Language} {Explanations}.
  In: Bengio, S., Wallach, H., Larochelle, H., Grauman, K., Cesa-Bianchi, N.,
  Garnett, R. (eds.) Advances in {Neural} {Information} {Processing} {Systems}.
  vol.~31. Curran Associates, Inc. (2018)

\bibitem{chauhan_topic_2022}
Chauhan, U., Shah, A.: Topic {Modeling} {Using} {Latent} {Dirichlet}
  allocation: {A} {Survey}. ACM Computing Surveys  \textbf{54}(7),  1--35 (Sep
  2022)

\bibitem{churchill_evolution_2022}
Churchill, R., Singh, L.: The {Evolution} of {Topic} {Modeling}. ACM Computing
  Surveys  \textbf{54}(10s),  1--35 (Jan 2022)

\bibitem{dennis_ai_2023}
Dennis, A.R., Lakhiwal, A., Sachdeva, A.: {AI} {Agents} as {Team} {Members}:
  {Effects} on {Satisfaction}, {Conflict}, {Trustworthiness}, and {Willingness}
  to {Work} {With}. Journal of Management Information Systems  \textbf{40}(2),
  307--337 (Apr 2023)

\bibitem{devlin_bert_2018}
Devlin, J., Chang, M.W., Lee, K., Toutanova, K.: {BERT}: {Pre}-training of
  {Deep} {Bidirectional} {Transformers} for {Language} {Understanding}  (2018)

\bibitem{dieng_topic_2020}
Dieng, A.B., Ruiz, F.J.R., Blei, D.M.: Topic {Modeling} in {Embedding}
  {Spaces}. Transactions of the Association for Computational Linguistics
  \textbf{8},  439--453 (Dec 2020)

\bibitem{greene_practical_2006}
Greene, D., Cunningham, P.: Practical solutions to the problem of diagonal
  dominance in kernel document clustering. In: Proceedings of the 23rd
  international conference on {Machine} learning - {ICML} '06. pp. 377--384.
  ACM Press, Pittsburgh, Pennsylvania (2006)

\bibitem{grootendorst_bertopic_2022}
Grootendorst, M.: {BERTopic}: {Neural} topic modeling with a class-based
  {TF}-{IDF} procedure  (2022)

\bibitem{hughes_ai_2025}
Hughes, L., Dwivedi, Y.K., Malik, T., Shawosh, M., Albashrawi, M.A., Jeon, I.,
  Dutot, V., Appanderanda, M., Crick, T., De’, R., Fenwick, M., Gunaratnege,
  S.M., Jurcys, P., Kar, A.K., Kshetri, N., Li, K., Mutasa, S., Samothrakis,
  S., Wade, M., Walton, P.: {AI} {Agents} and {Agentic} {Systems}: {A}
  {Multi}-{Expert} {Analysis}. Journal of Computer Information Systems pp.
  1--29 (Apr 2025)

\bibitem{mikolov_efficient_2013}
Mikolov, T., Chen, K., Corrado, G., Dean, J.: Efficient {Estimation} of {Word}
  {Representations} in {Vector} {Space}  (2013)

\bibitem{mittelstadt_explaining_2019}
Mittelstadt, B., Russell, C., Wachter, S.: Explaining {Explanations} in {AI}.
  In: Proceedings of the {Conference} on {Fairness}, {Accountability}, and
  {Transparency}. pp. 279--288. ACM, Atlanta GA USA (Jan 2019)

\bibitem{mohammad_large_2023}
Mohammad, A.F., Clark, B., Hegde, R.: Large {Language} {Model} ({LLM}) \&
  {GPT}, {A} {Monolithic} {Study} in {Generative} {AI}. In: 2023 {Congress} in
  {Computer} {Science}, {Computer} {Engineering}, \&amp; {Applied} {Computing}
  ({CSCE}). pp. 383--388. IEEE, Las Vegas, NV, USA (Jul 2023)

\bibitem{myers_qualitative_1997}
Myers, M.D.: Qualitative {Research} in {Information} {Systems}. MIS Quarterly
  \textbf{21}(2), ~241 (Jun 1997)

\bibitem{paullada_data_2021}
Paullada, A., Raji, I.D., Bender, E.M., Denton, E., Hanna, A.: Data and its
  (dis)contents: {A} survey of dataset development and use in machine learning
  research. Patterns  \textbf{2}(11),  100336 (Nov 2021)

\bibitem{pennington_glove_2014}
Pennington, J., Socher, R., Manning, C.: Glove: {Global} {Vectors} for {Word}
  {Representation}. In: Proceedings of the 2014 {Conference} on {Empirical}
  {Methods} in {Natural} {Language} {Processing} ({EMNLP}). pp. 1532--1543.
  Association for Computational Linguistics, Doha, Qatar (2014)

\bibitem{rajendran_local_2024}
Rajendran, B., Vidya, C.G., Sanil, J., Asharaf, S.: A {Local} {Explainability}
  {Technique} for {Graph} {Neural} {Topic} {Models}. Human-Centric Intelligent
  Systems  \textbf{4}(1),  53--76 (Jan 2024)

\bibitem{ries_reproducibility_2024}
Ries, T., Van Dalen-Oskam, K., Offert, F.: Reproducibility and explainability
  in digital humanities. International Journal of Digital Humanities
  \textbf{6}(1), ~1--7 (Jan 2024)

\bibitem{roder_exploring_2015}
Röder, M., Both, A., Hinneburg, A.: Exploring the {Space} of {Topic}
  {Coherence} {Measures}. In: Proceedings of the {Eighth} {ACM} {International}
  {Conference} on {Web} {Search} and {Data} {Mining}. pp. 399--408. ACM,
  Shanghai China (Feb 2015)

\bibitem{singla_overcoming_2019}
Singla, A., Bertino, E., Verma, D.: Overcoming the {Lack} of {Labeled} {Data}:
  {Training} {Intrusion} {Detection} {Models} {Using} {Transfer} {Learning}.
  In: 2019 {IEEE} {International} {Conference} on {Smart} {Computing}
  ({SMARTCOMP}). pp. 69--74. IEEE (Jun 2019)

\bibitem{taggu_deep_2024}
Taggu, A., Dubey, C., Paul, R.: Deep {Learning}-{Driven} {Sentiment}
  {Analysis}: {Unlocking} {Insights} in {Topic}-{Specific} {Twitter}
  {Conversations}. In: Proceedings of the 2024 6th {Asia} {Conference} on
  {Machine} {Learning} and {Computing}. pp. 33--37. ACM, Bangkok Thailand (Jul
  2024)

\bibitem{van_der_velden_explainable_2022}
Van Der~Velden, B.H., Kuijf, H.J., Gilhuijs, K.G., Viergever, M.A.: Explainable
  artificial intelligence ({XAI}) in deep learning-based medical image
  analysis. Medical Image Analysis  \textbf{79},  102470 (Jul 2022)

\bibitem{xie_knowledge_enhanced_2024}
Xie, Q., Tiwari, P., Ananiadou, S.: Knowledge-{Enhanced} {Graph} {Topic}
  {Transformer} for {Explainable} {Biomedical} {Text} {Summarization}. IEEE
  Journal of Biomedical and Health Informatics  \textbf{28}(4),  1836--1847
  (Apr 2024)

\bibitem{zhu_graph_2023}
Zhu, B., Cai, Y., Ren, H.: Graph neural topic model with commonsense knowledge.
  Information Processing \& Management  \textbf{60}(2),  103215 (Mar 2023)

\end{thebibliography}
\end{document}